\documentclass[letterpaper]{article} % DO NOT CHANGE THIS
\usepackage{aaai2026}  % DO NOT CHANGE THIS
\usepackage{times}  % DO NOT CHANGE THIS
\usepackage{helvet}  % DO NOT CHANGE THIS
\usepackage{courier}  % DO NOT CHANGE THIS
\usepackage[hyphens]{url}  % DO NOT CHANGE THIS
\usepackage{graphicx} % DO NOT CHANGE THIS
\usepackage{natbib}  % DO NOT CHANGE THIS AND DO NOT ADD ANY OPTIONS TO IT
\usepackage{caption} % DO NOT CHANGE THIS AND DO NOT ADD ANY OPTIONS TO IT
\usepackage{algorithm}
\usepackage{algorithmic}
\usepackage{booktabs}
\usepackage{multirow}
\usepackage{amsfonts}
\usepackage{amsmath}
\usepackage{fontawesome5}
\usepackage{xcolor}
\usepackage{nicematrix}
\usepackage{tikz}
\usepackage{newfloat}
\usepackage{listings}
\DeclareCaptionStyle{ruled}{labelfont=normalfont,labelsep=colon,strut=off} % DO NOT CHANGE THIS
\floatstyle{ruled}
\newfloat{listing}{tb}{lst}{}
\floatname{listing}{Listing}
\newcommand{\datasetone}{UPMC-Food101-CMML}
\newcommand{\datasettwo}{MM-IMDb-CMML}

\definecolor{baselinecolor}{gray}{.9}
\definecolor{demphcolor}{RGB}{144,144,144}

\usepackage[table]{xcolor}

\title{DeLo: Dual Decomposed Low-Rank Experts Collaboration for \\ Continual Missing Modality Learning}
\author{
    Xiwei Liu\textsuperscript{\rm 1}, Yulong Li\textsuperscript{\rm 1}, Feilong Tang\textsuperscript{\rm 1,2}, Imran Razzak\textsuperscript{\rm 1,}\footnote{Corresponding author}
}
\affiliations{
    \textsuperscript{\rm 1}Mohamed bin Zayed University of Artificial Intelligence \\
    \textsuperscript{\rm 2}Monash University\\

    \{Xiwei.Liu, Imran.Razzak\}@mbzuai.ac.ae
}

\usepackage{bibentry}
\begin{document}

\maketitle

\begin{abstract}
Adapting Large Multimodal Models (LMMs) to real-world scenarios poses the dual challenges of learning from sequential data streams while handling frequent modality incompleteness, a task known as Continual Missing Modality Learning (CMML). However, existing works on CMML have predominantly relied on prompt tuning, a technique that struggles with this task due to cross-task interference between its learnable prompts in their shared embedding space. A naive application of Low-Rank Adaptation (LoRA) with modality-shared module will also suffer modality interference from competing gradients. To this end, we propose DeLo, the first framework to leverage a novel dual-decomposed low-rank expert architecture for CMML. Specifically, this architecture resolves modality interference through decomposed LoRA expert, dynamically composing LoRA update matrix with rank-one factors from disentangled modality-specific factor pools. Embedded within a task-partitioned framework that structurally prevents catastrophic forgetting, this expert system is supported by two key mechanisms: a Cross-Modal Guided Routing strategy to handle incomplete data and a Task-Key Memory for efficient, task-agnostic inference. Extensive experiments on established CMML benchmarks demonstrate that our method significantly outperforms state-of-the-art approaches. This highlights the value of a principled, architecturally-aware LoRA design for real-world multimodal challenges. 
\end{abstract}

\begin{links}
    \link{Code}{https://github.com/Xiwei-web/DeLo}
    % \link{Datasets}{https://aaai.org/example/datasets}
    % \link{Extended version}{https://aaai.org/example/extended-version}
\end{links}

\section{Introduction}
Pre-trained multimodal models have shown great potential in many applications~\cite{radford2021learning, li2023blip}. When fine-tuning these pre-trained models on downstream tasks, missing modality issues often occur due to equipment failure, data corruption, privacy concerns, etc. Existing methods~\cite{zhao2021missing, ma2021smil, lee2023multimodal, jang2024towards} address missing modality issues by reconstructing missing information or aligning multimodal features. However, the sequential data in real-world applications will make these methods suffer from catastrophic forgetting~\cite{guo2025efficient}, leading to performance degradation. To handle forgetting, an intuitive idea is to store and retrain all old data but it creates large storage overheads and potential privacy-leak issues.

\begin{figure}[t] 
    \centering 
    \captionsetup{aboveskip=-2pt} 
    \includegraphics[width=\linewidth]{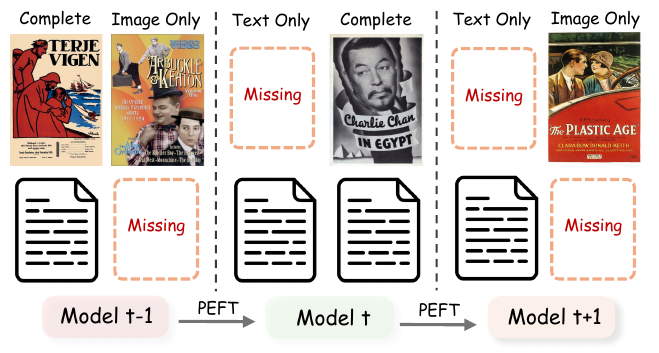} 
    \caption{Illustration of continual missing modality learning.}
    \label{fig:cmml} 
    \vspace{-1.5em}
\end{figure}

In recent years, continual learning (CL) has made great progress, such as replay-based methods~\cite{rolnick2019experience, buzzega2020dark, cha2021co2l}, regularization-based methods~\cite{kirkpatrick2017overcoming, zenke2017continual, aljundi2018memory}, and architecture-based methods~\cite{serra2018overcoming, li2019learn, ebrahimi2020adversarial}. Prompt-based methods~\cite{wang2022learning, wang2022dualprompt, wang2022s, smith2023coda} have attracted much attention due to their simplicity and effectiveness. However, most of these methods are unimodal and difficult to transfer to multimodal field. Multimodal methods~\cite{wang2022s, qian2023decouple} always depend on language-image models such as CLIP~\cite{radford2021learning}, which makes them difficult to apply to other fields where there are more modalities. LoRA-based methods~\cite{wistuba2023continual, liang2024inflora, chen2024dual} have also emerged. These approaches aim to better resolve the interference between new and old tasks to achieve a superior stability-plasticity trade-off.

Prompt-based methods were recently extended to address a more complex challenge: Continual Missing Modality Learning (Figure~\ref{fig:cmml})~\cite{zhao2024reconstruct, yue2025pal, guo2025efficient}. However, while this line of research has proven fruitful, its foundational reliance on prompt tuning as the parameter efficient fine tuning (PEFT) mechanism warrants closer examination. Recent studies~\cite{wistuba2024choice} argue that this choice is often an "undefended and unablated decision," with extensive evidence demonstrating that LoRA~\cite{hu2022lora} is substantially more performant than prompt tuning for CL. The core issue is the learnable prompts of prompt tuning are optimized within a shared embedding space, inherently limiting their capacity to prevent the cross-task interference that causes catastrophic forgetting. This evidence motivates a shift towards a LoRA-based approach. However, a naive application of standard LoRA for CMML is also insufficient. Such an approach would still suffer from modality interference, where conflicting updates from these heterogeneous data sources (e.g., image-only vs. text-only) would lead to the learning of entangled and suboptimal adaptations. Table~\ref{tab:method compare} highlights that there is a clear gap where no existing method effectively tailors a powerful PEFT like LoRA to the dual challenges of CMML.

Based on these findings, our work proposes DeLo, the first LoRA-based framework designed to tackle the CMML problem by leveraging a novel dual-\textbf{De}composed \textbf{Lo}w-rank expert architecture. DeLo adopts dynamic decomposed low-rank expert with factors selected from Modality-Specific Factor Pool through instance-based Cross-Modal Guided Routing strategy. DeLo features a task-partitioned expert architecture to physically isolate knowledge, and utilizes a Task-Key Memory to enable task-agnostic inference.
\begin{table}[t] 
    % \captionsetup{belowskip=1pt}
    \centering 
    \resizebox{\linewidth}{!}{
    \begin{tabular}{@{}l|cccc@{}}
    \toprule
       \multirow{2}{*}{Method} &  Missing & Continual &  Prompt  & \multirow{2}{*}{LoRA} \\
        & Modality & Learning &  Tuning &  \\\midrule
     MAP~\cite{lee2023multimodal} &  \faCheck & \faTimes & \faCheck  & \faTimes \\
     MSP~\cite{jang2024towards} & \faCheck & \faTimes & \faCheck  & \faTimes \\
     L2P~\cite{wang2022learning} &  \faTimes & \faCheck & \faCheck  & \faTimes \\
     DualPrompt~\cite{wang2022dualprompt} &  \faTimes & \faCheck & \faCheck  & \faTimes \\
     InfLoRA~\cite{liang2024inflora} &  \faTimes & \faCheck & \faTimes  & \faCheck \\
     RebQ~\cite{zhao2024reconstruct} &  \faCheck & \faCheck & \faCheck  & \faTimes \\
     \midrule
     DeLo (Ours) &  \textcolor{red}{\faCheck} & \textcolor{red}{\faCheck} & \textcolor{red}{\faTimes}  & \textcolor{red}{\faCheck} \\
    \bottomrule
    \end{tabular}
    }
    \caption{Comparison of related representative works by problem area and adaptation technique.}
    \vspace{-1em}
    \label{tab:method compare}
\end{table}
We conduct extensive experiments on two multimodal datasets: {\datasetone} and {\datasettwo}. Our proposed method can consistently outperform baselines and state-of-the-art methods significantly on both two datasets. 
% Besides, the number of trainable parameters only accounts for 5-7$\%$ of the parameters of the backbone network, indicating our method is parameter-efficient. 

Our main contributions can be summarized as follows:
\begin{itemize}
    \item To the best of our knowledge, we are the first to propose a LoRA-based framework for the Continual Missing Modality Learning task, bridging a significant gap in this commonly encountered real-world scenario.

    \item We propose a novel Dual Decomposed Low-Rank Expert architecture that simultaneously addresses catastrophic forgetting and modality interference. The LoRA experts is composed of selectable low-rank factors from disentangled modality-specific factor pools.

    \item We demonstrate through extensive experiments that DeLo achieves state-of-the-art performance on CMML benchmarks, delivering superior average performance while structurally mitigating catastrophic forgetting.
\end{itemize}

\begin{figure*}[t] 
    \centering 
    \captionsetup{aboveskip=-34pt} 
    \includegraphics[width=\textwidth]{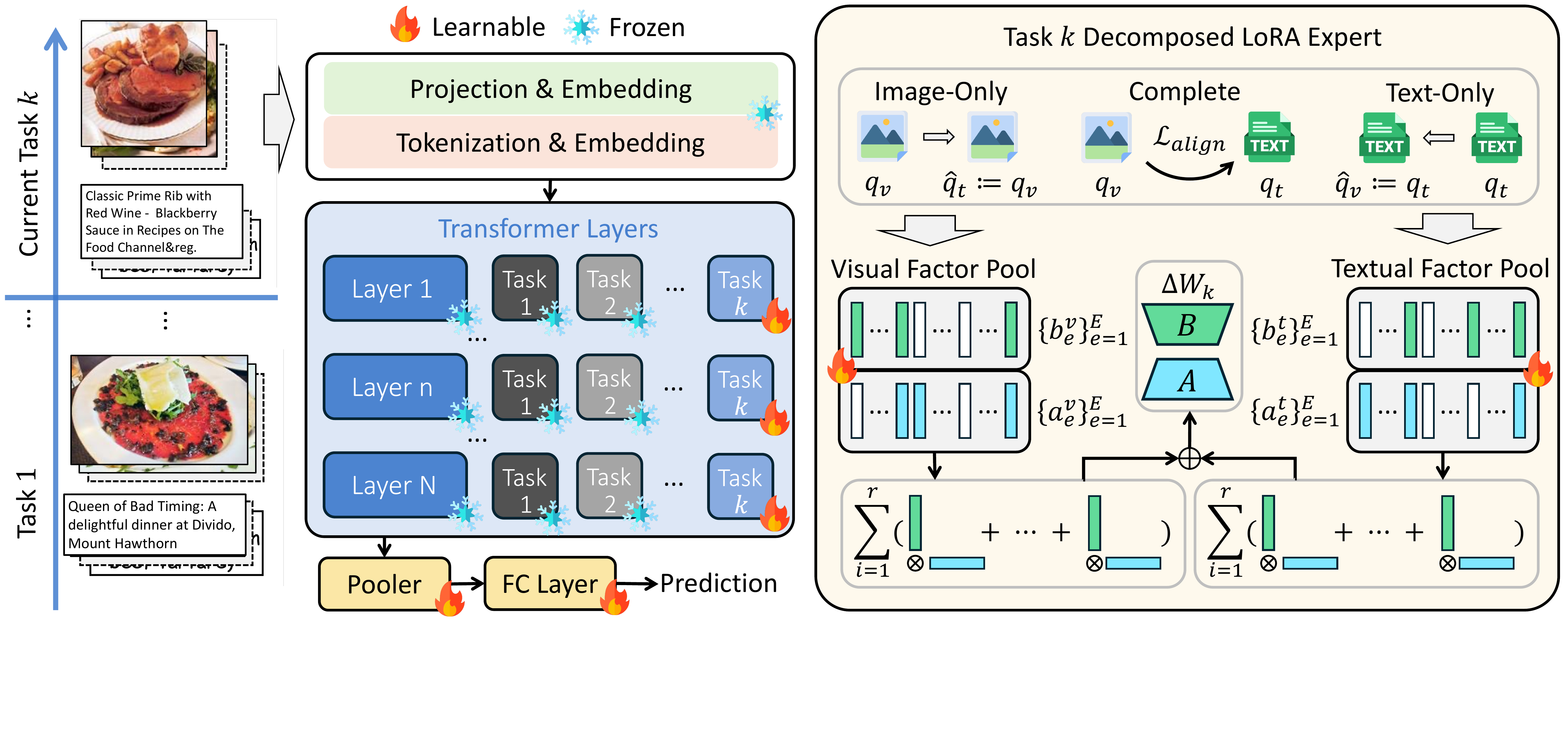} 
    \caption{\textbf{Overview of our proposed DeLo framework.} The left panel illustrates the task-partitioned architecture for continual learning. To prevent catastrophic forgetting, only the expert modules and classifier for the current task $k$ are trainable, while the main backbone and all modules for previous tasks (1 to $k-1$) are frozen. The right panel details the dual decomposed LoRA expert for a given Task $k$, which consists of Modality-Specific Factor Pools (Visual: $\mathcal{P}_k^{\mathrm{v}}=\{(a_e^{\mathrm{v}}, b_e^{\mathrm{v}})\}_{e=1}^E$ and Textual: $\mathcal{P}_k^{\mathrm{t}}=\{(a_e^{\mathrm{t}}, b_e^{\mathrm{t}})\}_{e=1}^E$). For each input, factors are dynamically selected from these pools to compose modality-specific weight adjustments $\Delta W^{\mathrm{v}} = \sum_{i=1}^{r} b_i^{\mathrm{v}} \otimes a_i^{\mathrm{v}}$ and  
    $\Delta W^{\mathrm{t}} = \sum_{j=1}^{r} b_j^{\mathrm{t}} \otimes a_j^{\mathrm{t}}$, whose sum forms the final update $\Delta W_k$. The top of this panel shows our Cross-Modal Guided Routing for handling missing modalities, where the query from an available modality serves as a proxy for the missing one (e.g., $\hat{q}_{\mathrm{t}}:=q_{\mathrm{v}}$ for image-only input) and the Alignment Loss $\mathcal{L}_{\mathrm{align}}$ for complete data.}
    \label{fig:methdology} 
\end{figure*}

\section{Related Work}
\subsection{Multi-Modal Learning with Missing Modalities}
The challenge of handling missing modalities is a significant bottleneck in multimodal learning, often causing severe performance degradation~\cite{lee2023multimodal, jin2023rethinking}. Research to mitigate this issue has broadly followed three main directions. One line of work focuses on generative solutions~\cite{cai2018deep, zhang2024unified, ke2025knowledge}, such as directly synthesizing the data of the absent modality using the available information. Other approaches operate in the feature space, aiming to learn robust representations that are resilient to missing data. These include methods that align features by translating between modalities~\cite{pham2019found}, predict the missing modality's representation from available ones~\cite{zhao2021missing}, or estimate latent features via Bayesian meta-learning~\cite{ma2021smil}.

More recently, a popular parameter-efficient strategy has emerged that leverages learnable prompt tokens within pre-trained transformers to act as proxy representation for missing information~\cite{lee2023multimodal, woo2023towards, jang2024towards, zhao2024reconstruct, yue2025pal, guo2025efficient}. However, these prompt-based methods, much like the other strategies discussed, are predominantly designed for static fixed datasets. This shared limitation makes them ill-equipped for real-world scenarios where data arrive sequentially, a core challenge addressed by CMML.

\subsection{Continual Learning}
Continual learning (CL) endeavors to empower models to learn from a sequential stream of tasks. The primary barrier to this goal is catastrophic forgetting~\cite{mccloskey1989catastrophic}, a phenomenon in which acquiring new knowledge leads to the abrupt erasure of previously learned information. Traditional CL strategies, which typically learn from scratch, can be categorized into three families: (1) Regularization-based methods~\cite{zenke2017continual, aljundi2018memory, jung2020continual} protect prior knowledge by imposing constraints on updates to parameters deemed critical for past tasks. (2) Replay-based methods~\cite{aljundi2019online, chrysakis2020online, liang2023loss, liang2023adaptive} store a small buffer of representative samples from previous tasks to rehearse during new task training. (3) Architecture-based methods~\cite{li2019learn, hung2019compacting, ebrahimi2020adversarial} mitigate interference by dynamically expanding the network or isolating subsets of task-specific parameters. Although effective to some extent, these methods often face scalability challenges as the complexity of the model or the number of tasks increases.

The paradigm has recently shifted toward leveraging large pre-trained models~\cite{liu2025towards, he2022masked} for continual learning. Early attempts to fully fine-tune these large models proved to be inefficient in parameters~\cite{boschini2022transfer, zheng2023preventing}. This has spurred the exploration of PEFT for CL. Prominent examples include prompt adjustment~\cite{wang2022learning, wang2022dualprompt, khan2023introducing, smith2023coda}, which has demonstrated strong performance, particularly in class-incremental scenarios, and various methods based on LoRA~\cite{wistuba2023continual, liang2024inflora, chen2024dual},  which focus on mitigating intertask interference during sequential learning. However, effective application of these powerful PEFT strategies, especially LoRA, to continual learning tasks with incomplete data remains a key open challenge.

\section{Proposed Method}
\subsection{Problem Formulation}
Our work tackles the challenging yet practical scenario of CMML, where a model must sequentially learn from tasks containing potentially incomplete multimodal data. 
This setting imposes three real-world constraints: (1) parameter efficiency for on-device adaptation, since full fine-tuning of large pre-trained models is infeasible; 
(2) privacy and storage limitations that prohibit data replay, making the model prone to catastrophic forgetting as new tasks arrive; and 
(3) autonomous system must perform task-agnostic inference, enabling classification without prior task IDs information.

% Our work addresses the challenging yet practical scenario of CMML, where a model must sequentially learn from a stream of tasks, each containing potentially incomplete multimodal data. 
% % CMML imposes three real-world constraints: (1) parameter efficiency for on-device learning, (2) no data replay to preserve privacy, and (3) task-agnostic inference without prior task IDs.
% This paradigm is defined by three critical constraints that mirror real-world deployment challenges. First, the necessity for on-device applications demands a parameter-efficient adaptation strategy, as fine-tuning the entire large pre-trained model is infeasible. Second, privacy and storage considerations often prohibit the rehearsal of past data, exposing the model to the risk of catastrophic forgetting, where performance on older tasks degrades significantly as new ones are learned. Finally, a truly autonomous system must perform task-agnostic inference, classifying inputs without explicit knowledge of their originating task.

Formally, we consider a sequence of $T$ tasks, $\mathcal{T} = \{\mathcal{T}_1, \ldots, \mathcal{T}_T\}$. For any given task $\mathcal{T}_k$, the model has access only to its corresponding training dataset $\mathcal{D}_k$. To simplify but without loss of generality, we consider a multimodal dataset with text (t) and visual (v) modalities, where each task's dataset $\mathcal{D}_k$ is structured as a composite of three subsets, $\mathcal{D} = \{ \mathcal{D}^{\mathrm{t}}, \mathcal{D}^{\mathrm{v}}, \mathcal{D}^{\mathrm{c}}\}$, to capture the nature of modality-missing. 
% \begin{equation}
%     \mathcal{D} = \{ \mathcal{D}^{\mathrm{t}}, \mathcal{D}^{\mathrm{v}}, \mathcal{D}^{\mathrm{c}}\}.
% \end{equation}
Specifically, a set of modality-complete samples $\mathcal{D}^{\mathrm{c}} = \{(\mathbf{t}_i, \mathbf{v}_i, \mathbf{y}_i)\}_{i=1}^{N_{\mathrm{c}}}$; a text-only set $\mathcal{D}^{\mathrm{t}} = \{(\mathbf{t}_i, \mathbf{y}_i)\}_{i=1}^{N_{\mathrm{t}}}$; and a visual-only set $\mathcal{D}^{\mathrm{v}} = \{(\mathbf{v}_i, \mathbf{y}_i)\}_{i=1}^{N_{\mathrm{v}}}$, where $N_{\mathrm{c}}, N_{\mathrm{t}}$ and $N_{\mathrm{v}}$ are the number of samples in each respective subset. Here, $\mathbf{t}_i/\mathbf{v}_i$ denotes the embedding of textual/visual content and $\mathbf{y}_i \in \mathbb{R}^{C}$ is the label vector where $C$ is the number of classes. Following~\cite{lee2023multimodal}, we assign dummy inputs $\tilde{\mathbf{t}}/\tilde{\mathbf{v}}$ to represent missing-modality data and obtain $\tilde{\mathcal{D}}^{\mathrm{t}}= \{(\mathbf{t}_{i}, \tilde{\mathbf{v}}, \mathbf{y}_{i})\}_{i=1}^{N_{\mathrm{t}}}$ and $ \tilde{\mathcal{D}}^{\mathrm{v}} = \{(\tilde{\mathbf{t}}, \mathbf{v}_{i}, \mathbf{y}_{i})\}_{i=1}^{N_{\mathrm{v}}}$. During the training, data from previous tasks is not available. In the inference phase, the task ID is unknown.

% \vspace{-1em}
% \subsection{Overview}
\subsection{Decomposed Low-Rank Expert Architecture}

\begin{figure}[t] 
    \centering 
    \captionsetup{aboveskip=-34pt} 
    \includegraphics[width=0.72\linewidth]{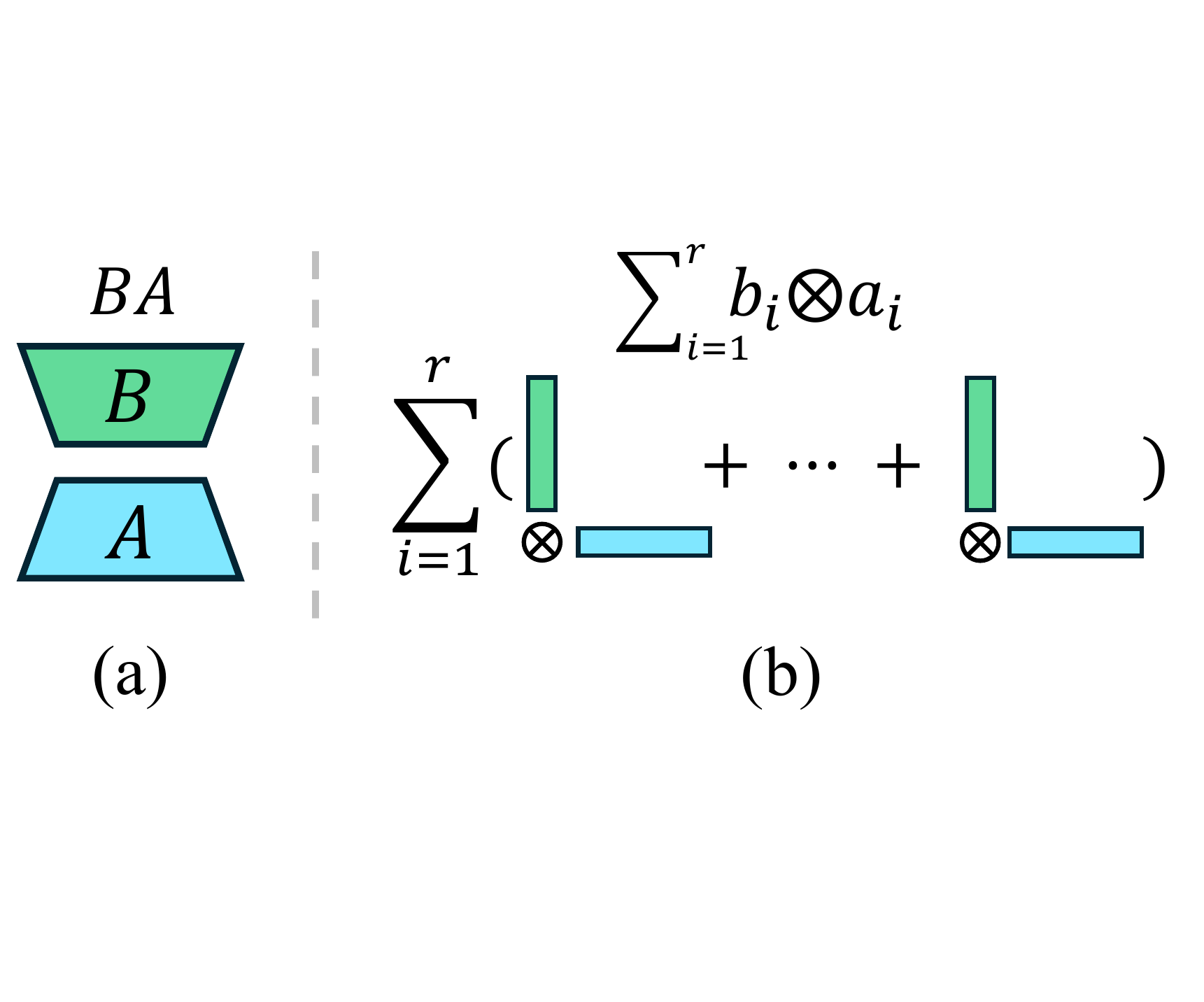} 
    \caption{Comparison of LoRA parameterizations. (a) Conventional LoRA, where the weight adjustment matrix is represented as a product of two low-rank matrices ($BA$). (b) Decomposed LoRA, where the adjustment is expressed as a sum of $r$ rank-one matrices, each formed by the outer product of a vector pair $\sum_{i=1}^r b_i \otimes a_i$.}
    \vspace{-1em}
    \label{fig:decomposed LoRA} 
\end{figure}

Inspired by the concept of Mixture-of-Experts~\cite{shazeer2017outrageously} and its application to dynamic adapter construction~\cite{shen2024multimodal}, DeLo leverages low-rank decomposition factors as a pool of selectable experts. In standard LoRA, a frozen pre-trained weight matrix $W \in \mathbb{R}^{d_{\text{out}} \times d_{\text{in}}}$ is augmented by injecting a parallel, trainable weight adjustment matrix $\Delta W \in \mathbb{R}^{d_{out} \times d_{in}}$, where $d_{in}$ and $d_{out}$ are the input and output dimensions respectively. This adjustment is efficiently parameterized by decomposing it into two smaller, low-rank matrices: $\Delta W = B A$, where $A \in \mathbb{R}^{r \times d_{\text{in}}}$ and $B \in \mathbb{R}^{d_{\text{out}} \times r}$. The rank $r$ is much smaller than the original dimensions ($r \ll \min(d_{\text{in}}, d_{\text{out}})$) to ensure high parameter efficiency. The forward pass of a LoRA-adapted layer computes the input hidden state $h$ as follows: 
\begin{equation}
    \tilde{h} = W h + \Delta W h= W h + \alpha \cdot B A h,
\end{equation}
with the $\alpha \geq 1$ controls the influence of the weight adjustment matrices. As shown in Figure~\ref{fig:decomposed LoRA}, DeLo departs from this static formulation by dynamically composing this weight adjustment matrix from a larger pool of rank-one factors. This can be expressed via tensor decomposition as:
\begin{align}
    \Delta W = BA = \sum_{i=1}^r b_i \otimes a_i,
    \label{eq:mixlora}
\end{align}
where $\otimes$ denotes the outer product and each $(a_i, b_i)$ is the selected vector pair with $a_i \in \mathbb{R}^{d_{in} \times 1}$ and $b_i \in \mathbb{R}^{d_{out} \times 1}$.  Instead of learning a single $B$ and $A$ matrix, DeLo maintains expanded pools of expert factors. These pools contain $E$ total pool pairs, $\{a_e\}_{e=1}^E, \{b_e\}_{e=1}^E, E > r$. This allows the model to dynamically construct a unique $\Delta W $ for specific inputs by selecting $r$ appropriate factors. Following LoRA~\cite{hu2022lora}, we use a random Gaussian initialization for $\{a_e\}_{e=1}^E$ and zero for $\{b_e\}_{e=1}^E$.

\subsection{Decomposed modality-specific factor pool}
A central challenge in adapting multimodal models is the potential for modality interference. When a single, unified pool of experts is used to learn from distinct data streams like vision and text, its parameters are updated by competing gradients, which can lead to the learning of entangled and suboptimal representations. Inspired by the knowledge decomposition strategy in RebQ~\cite{zhao2024reconstruct}, DeLo employs modality-specific factor pools to ensure the model learns specialized, disentangled adaptations for visual and textual features without interference. Specifically, the expert system is comprised of two distinct and independent sub-pools: a vision factor pool $\mathcal{P}^{\mathrm{v}}$ and a text factor pool $\mathcal{P}^{\mathrm{t}}$. The vision factor pool $\mathcal{P}^{\mathrm{v}}$ contains $E_{\mathrm{v}}$ pairs of learnable low-rank factors dedicated to the visual modality: $\mathcal{P}^{\mathrm{v}} = \{ (a_e^{\mathrm{v}}, b_e^{\mathrm{v}}) \}_{e=1}^{E_{\mathrm{v}}}$, where $a_e^{\mathrm{v}}$ and $b_e^{\mathrm{v}}$ are the expert factors for vision. Similarly, the text factor pool $\mathcal{P}^{\mathrm{t}}$ contains $E_{\mathrm{t}}$ pairs of factors for the textual modality: $\mathcal{P}^{\mathrm{t}} = \{ (a_e^{\mathrm{t}}, b_e^{\mathrm{t}}) \}_{e=1}^{E_{\mathrm{t}}}$.

We adopt the instance-based routing method from MixLoRA~\cite{shen2024multimodal} to select factors from expanded pools. For a given input hidden state $h \in \mathbb{R}^{\text{seq} \times d_{\text{in}}}$ from the preceding layer, a query signal $q = \text{Avg}(h)$ is first generated by averaging the hidden states across the sequence dimension. The input query $q$ is then processed through a dense layer with weights $W_A \in \mathbb{R}^{E \times d_{\text{in}}}$ and a softmax function to generate scores over the expert pool. The final selection vector $g_A \in \{0, 1\}^E$ is determined by a $\text{top}_r(\cdot)$ operation that creates a binary mask identifying the top-$r$ scoring experts:
\begin{equation}
    g_A = \text{top}_r(\text{softmax}(W_A \cdot q)),
    \label{Eq. Selection}
\end{equation}

The selection of factors for the B matrix employs a more comprehensive process to ensure a cohesive adaptation. It is guided by two signals: the same input query $q$, and the composition of the $A$ matrix that was just selected. The selection scores derived from both signals are fused via element-wise addition to determine the final probabilities for the $B$ factors. The resulting selection vector $g_B$, therefore identifies a set of $B$ factors that are not only relevant to the input but also highly compatible with the chosen $A$ factors.

We finally select $r$ vision experts from $\mathcal{P}^{\mathrm{v}}$ to compose the vision-specific adjustment matrix $\Delta W^{\mathrm{v}}$. Likewise, it selects $r$ text experts from $\mathcal{P}^{\mathrm{t}}$ to compose the text-specific adjustment matrix $\Delta W^{\mathrm{t}}$. These are formulated as:
\begin{equation}
    \Delta W^{\mathrm{v}} = \sum_{i=1}^{r} b_i^{\mathrm{v}} \otimes a_i^{\mathrm{v}} \quad \text{and} \quad 
    \Delta W^{\mathrm{t}} = \sum_{j=1}^{r} b_j^{\mathrm{t}} \otimes a_j^{\mathrm{t}}.
\end{equation}
The final weight adjustment matrix for the input is the collaborative sum of these two modality-specific adjustments. The complete forward pass for a layer within the model is therefore: $\tilde{h} = Wh + (\Delta W^{\mathrm{v}} + \Delta W^{\mathrm{t}}) h$.

\subsection{Cross-Modal Guided Routing}
Our method presents a unique challenge: \textit{How does the system perform modality-specific factor selection when one modality is missing?} We address this by using the query from the available modality as a proxy for the missing one, leveraging the shared semantic space of the multimodal backbone. For a complete sample with visual and textual hidden states $h_{\mathrm{v}}$ and $h_{\mathrm{t}}$, their averaged queries $q_{\mathrm{v}}=\text{Avg}(h_{\mathrm{v}})$ and $q_{\mathrm{t}}=\text{Avg}(h_{\mathrm{t}})$ are semantically aligned ($q_{\mathrm{v}}\!\approx\!q_{\mathrm{t}}$). Thus, when text is missing, the proxy query is simply $\hat{q}_{\mathrm{t}}:=q_{\mathrm{v}}$. Expert selection for the missing modality then follows the standard routing rule:
% To address this, we use the query signal from the available modality as a proxy to guide the factor selection process for the missing modality. The feasibility of this approach stems from the shared semantic space learned by the multimodal backbone, which aligns the representations of corresponding concepts. This alignment ensures that for a complete data pair with visual hidden states $h_{\mathrm{v}}$ and textual hidden states $h_{\mathrm{t}}$, their respective query signals $q_{\mathrm{v}} = \text{Avg}(h_{\mathrm{v}})$ and $q_{\mathrm{t}} = \text{Avg}(h_{\mathrm{t}})$ are semantically approximate ($q_{\mathrm{v}} \approx q_{\mathrm{t}}$). Therefore, when a modality is missing, such as text, we use the query from the available visual modality as its high-fidelity semantic proxy. The proxy query for the missing text modality $\hat{q}_t$ is formally defined as: $\hat{q}_{\mathrm{t}} := q_{\mathrm{v}}$. The selection of text-specific experts then proceeds by substituting this proxy query into the standard selection Equation~\ref{Eq. Selection} (for matrix $A$, as an example):
\begin{equation}
    g_A^{\mathrm{t}} = \text{top}_r(\text{softmax}(W_A^{\mathrm{t}} \cdot \hat{q}_{\mathrm{t}})).
\end{equation}
Because both queries share the same hidden dimension $d_{\text{in}}$, this substitution requires no extra projection, keeping the routing mechanism simple and seamless.

% This direct substitution is technically seamless, as the uniform hidden dimension $d_{\text{in}}$ of the Transformer backbone ensures that $q_{\mathrm{v}}$ and $q_{\mathrm{t}}$ are dimensionally identical, requiring no additional projection layers. This format and dimension consistency allows the query signal from one modality to be directly passed to the routing weight matrix of the other without requiring any projection or modification, making the mechanism both efficient and elegant.

\begin{table*}[t]
    % \label{tab:food101}
    \centering
    \resizebox{\linewidth}{!}{
    \begin{tabular}{@{}ccc|cc|cc|cc|cc|cc|cc@{}}
    \toprule
    \multirow{2}{*}{$\eta$} & \multirow{2}{*}{\#Image} & \multirow{2}{*}{\#Text} & \multicolumn{2}{c|}{MAP} & \multicolumn{2}{c|}{MSP} & \multicolumn{2}{c|}{L2P} & \multicolumn{2}{c|}{DualPrompt} & \multicolumn{2}{c|}{RebQ} & \multicolumn{2}{c}{DeLo} \\
     &  &  & AP~($\uparrow$) & FG~($\downarrow$) & AP~($\uparrow$) & FG~($\downarrow$) & AP~($\uparrow$) & FG~($\downarrow$)  & AP~($\uparrow$) & FG~($\downarrow$) & AP~($\uparrow$) & FG~($\downarrow$) & AP~($\uparrow$) & FG~($\downarrow$)\\ \midrule
      % \rowcolor[gray]{0.9} \multicolumn{15}{c}{\textbf{{\datasetone}}}\\
       % \midrule
    \multirow{3}{*}{$10\%$} & $100\%$ & $90\%$  & 20.66 & 82.50 & 21.45 & 80.03 & 34.09 & 5.80 & 59.56 & 3.38 & 68.67 & 9.50  & \textbf{77.19} & 10.21\\ 
                             & $90\%$ & $100\%$ & 21.53 & 82.20 & 23.29 & 78.98 & 35.21 & 5.51 & 59.90 & 1.72 &  72.46 & 7.64  & \textbf{80.23} & 7.88 \\ 
                             & $95\%$ & $95\%$  & 22.84 & 80.13 & 22.12 & 79.35 & 34.00 & 6.55 & 58.18 & 6.32 &  71.06 & 8.22  & \textbf{78.12}  & 10.32\\ \midrule
    \multirow{3}{*}{$30\%$} & $100\%$ & $70\%$ & 16.57 & 83.77 & 21.37 & 79.95 & 30.40 & 6.53 & 51.86 & 6.46 &  62.06 & 10.76  & \textbf{74.08}  & 9.69\\ 
                             & $70\%$ & $100\%$ & 24.00 & 77.12 & 22.21 & 76.02 & 34.75 & 4.66 & 52.66 & 5.11 & 71.62 & 6.44  & \textbf{78.20}  & 6.34\\ 
                             & $85\%$ & $85\%$  & 20.66 & 79.81 & 21.89 & 79.01 & 29.76 & 6.87 & 48.56 & 8.73 & 66.37 & 8.07  &  \textbf{73.63}& 8.89 \\ \midrule
    \multirow{3}{*}{$50\%$} & $100\%$ & $50\%$ & 18.18 & 79.77 & 17.88 & 77.31 & 28.95 & 6.33 & 47.70 & 6.09 &  55.87 & 12.07  & \textbf{71.41} & 14.77\\ 
                             & $50\%$ & $100\%$ & 23.85 & 75.59 & 21.57 & 77.99 & 32.18 & 4.55 & 50.22 & 5.03 & 69.23 & 5.84  & \textbf{76.38} & 7.39\\ 
                             & $75\%$ & $75\%$  & 18.66 & 79.80 & 18.10 & 78.04 & 25.30 & 6.17 & 43.67 & 8.52 & 62.40 & 8.24 & \textbf{70.59} & 8.27\\ \midrule
    \multirow{3}{*}{$70\%$} & $100\%$ & $30\%$  & 17.68 & 77.82 & 19.29 & 78.17 & 26.57 & 5.80 & 43.09 & 8.96 &  50.00 & 12.47  & \textbf{70.04} & 14.91\\ 
                             & $30\%$ & $100\%$ & 22.48 & 75.21 & 21.22 & 75.21 & 30.43 & 3.99 & 50.28 & 5.19 & 69.41 & 3.73  & \textbf{74.80} & 5.32\\ 
                             & $65\%$ & $65\%$  & 20.00 & 75.95 & 19.76 & 76.54 & 24.62 & 5.66 & 40.69 & 7.26 & 59.92 & 8.56 & \textbf{67.42} & 9.41\\ \midrule
    \multirow{3}{*}{$90\%$} & $100\%$ & $10\%$  & 16.92 & 76.60 & 18.62 & 75.14 & 24.94 & 5.60 & 37.26 & 11.46 & 48.15 & 12.76 & \textbf{67.30} & 15.68 \\ 
                             & $10\%$ & $100\%$ & 24.89 & 70.84 & 21.69 & 71.23 & 29.77 & 4.54 & 51.16 & 4.80 &  67.71 & 4.71 &  \textbf{74.21} & 4.83\\ 
                             & $55\%$ & $55\%$ & 18.41 & 75.19 & 20.45 & 74.31 & 24.62 & 5.66 & 40.69 & 7.29 & 54.67 & 8.78  & \textbf{63.37}  & 10.97\\

    \bottomrule
    \end{tabular}
    }
    % \captionsetup{belowskip=1pt}
    \caption{\textbf{Performance comparison on {\datasetone} dataset}. We report AP and FG  based on classification accuracy. Missing Modality can occur in both training and testing. The best performance is in \textbf{bold}.}
     \label{tab:food101}
\end{table*}
\subsection{Task-Key Routing for Task-Agnostic Inference}
To prevent catastrophic forgetting, DeLo employs a task-partitioned architecture. The complete set of expert pools $\mathcal{P}$ can be defined as a collection of $T$ task-specific pools:
\begin{equation}
\mathcal{P} = \{(\mathcal{P}_1^{\mathrm{v}}, \mathcal{P}_1^{\mathrm{t}}), (\mathcal{P}_2^{\mathrm{v}}, \mathcal{P}_2^{\mathrm{t}}), ..., (\mathcal{P}_T^{\mathrm{v}}, \mathcal{P}_T^{\mathrm{t}})\},
\end{equation}
During the training of task $\mathcal{T}_k$ , only the parameters within its corresponding pool $\mathcal{P}_k$ and the classifier head are updated, while all previously learned pools and heads are frozen.

This partitioned design necessitates a mechanism for task-agnostic inference to select the correct expert pool for a given input. To this end, we introduce a lightweight, non-trainable buffer designed to dynamically route inputs without explicit task IDs. This module maintains a set of representative vectors $\{\text{key}_k\}_{k=1}^T$, where each $\text{key}_k \in \mathbb{R}^{d_{in}}$ serves as a unique centroid for task $\mathcal{T}_k$. During the training of task $\mathcal{T}_k$, its key is updated smoothly using the exponential moving average (EMA). At each training step $t$, the key is updated using the average query signal from $q_{\text{batch}}$ the current batch as follows:
\begin{equation}
\text{key}_{k,t} = \beta \cdot \text{key}_{k, t-1} + (1-\beta) \cdot q_{\text{batch}, t},
\end{equation}
where $\beta$ is the EMA momentum parameter, and $\text{key}_{k,t-1}$ is the key's state before update. At inference, given a test input with query signal $q_{\text{test}}$, the module predicts the task ID $k^{\star}$ by finding the key with the highest cosine similarity:
\begin{equation}
k^{\star} = \underset{k \in {1, ..., T}}{\text{argmax}} \left( \text{cos}(q_{\text{test}}, \text{key}_k) \right).
\end{equation}
The corresponding pool $\mathcal{P}_{k^*}$ is then activated to process the input. This approach is effective as each Task-Key serves as a robust representation of its task's feature cluster.

\subsection{Training Objective}
In this work, 
we address the classification task and adopt cross-entropy~(CE) loss $\mathcal{L}_{\mathrm{c}}$ (binary cross-entropy loss for the multi-label classification task) for training. To mitigate the potential modality bias inherent in our cross-modal guided routing mechanism when a modality is missing, we introduce two auxiliary losses computed exclusively on the modality-complete samples within each batch. An Alignment Loss $\mathcal{L}_{\mathrm{align}}$ explicitly encourages the query signals from paired visual ($q_{\mathrm{v}}$) and textual ($q_{\mathrm{t}}$) data to inhabit a shared semantic space by minimizing their cosine distance: 
\begin{equation}
    \mathcal{L}_{\mathrm{align}} = 1-\mathrm{cos}(q_{\mathrm{v}}, q_{\mathrm{t}}),
\end{equation}
thereby enhancing the quality of the proxy signal. Furthermore, to ensure the proxy signal elicits a similar reasoning process as the authentic one, we employ a Consistency Loss $\mathcal{L}_{\mathrm{con}}$. This term minimizes the KL divergence between the model's output distributions when a router is guided by the true query versus the proxy query. The final objective is:
\begin{equation}
    \mathcal{L} = \mathcal{L}_{\mathrm{c}} + \lambda_1 \mathcal{L}_{\mathrm{align}} + \lambda_2 \mathcal{L}_{\mathrm{con}}.
\end{equation}
where $\lambda_1$ and $\lambda_2$ are hyperparameters that balance the regularization. This joint optimization forces the model not only to be accurate but also to learn robust and consistent cross-modal representations, which are crucial for effective performance in modality-missing scenarios.

\begin{table*}[t]
    % \label{tab:mmimdb}
    \centering
    \resizebox{\linewidth}{!}{
    \begin{tabular}{@{}ccc|cc|cc|cc|cc|cc|cc@{}}
    \toprule
    \multirow{2}{*}{$\eta$} & \multirow{2}{*}{\#Image} & \multirow{2}{*}{\#Text} & \multicolumn{2}{c|}{MAP} & \multicolumn{2}{c|}{MSP} & \multicolumn{2}{c|}{L2P} & \multicolumn{2}{c|}{DualPrompt} & \multicolumn{2}{c}{RebQ} & \multicolumn{2}{c}{DeLo}\\
     &  &  & AP~($\uparrow$) & FG~($\downarrow$) & AP~($\uparrow$) & FG~($\downarrow$) & AP~($\uparrow$) & FG~($\downarrow$)  & AP~($\uparrow$) & FG~($\downarrow$) & AP~($\uparrow$) & FG~($\downarrow$) & AP~($\uparrow$) & FG~($\downarrow$)\\ \midrule
    \multirow{3}{*}{$10\%$} & $100\%$ & $90\%$  & 15.77 & 34.67 & 19.49 & 35.11 & 13.67 & 8.96 & 22.79 & 20.42 & 24.19 & 32.46 &  \textbf{29.14}    & 32.11 \\ 
                             & $90\%$ & $100\%$ & 16.63 & 33.30 & 19.01 & 34.89 & 13.54 & 8.26 & 24.82 & 24.09 & 27.35 & 28.88 &  \textbf{34.77}    & 30.36 \\ 
                             & $95\%$ & $95\%$  & 17.24 & 31.28 & 15.69 & 32.17 & 13.90 & 8.42 & 24.19 & 24.04 & 26.90 & 28.26 &   \textbf{31.84}   & 30.72\\ \midrule
    \multirow{3}{*}{$30\%$} & $100\%$ & $70\%$ & 19.85 & 32.14 & 20.01 & 34.87 & 10.68 & 8.03 & 22.96 & 23.96 & 24.75 & 25.86 &   \textbf{29.28}   & 32.79\\ 
                             & $70\%$ & $100\%$ & 17.19 & 31.09 & 18.31 & 32.45 & 10.98 & 7.27 & 20.98 & 19.67 & 26.68 & 27.89  &  \textbf{34.56}    & 32.38\\ 
                             & $85\%$ & $85\%$  & 14.44 & 33.29 & 14.20 & 35.57 & 11.58 & 7.90 & 22.18 & 20.72 & 25.80 & 26.95  &  \textbf{30.62}    & 29.23\\ \midrule
    \multirow{3}{*}{$50\%$} & $100\%$ & $50\%$  & 19.19 & 36.18 & 15.93 & 35.29 & 9.54 & 5.54 & 20.04 & 21.48 & 23.89 & 25.27  &   \textbf{28.88}   & 33.18\\ 
                             & $50\%$ & $100\%$ & 18.83 & 32.02 & 16.78 & 34.14 & 9.77 & 4.88 & 19.97 & 20.24 & 26.45 & 27.53  &  \textbf{32.83}    & 31.09\\ 
                             & $75\%$ & $75\%$  & 17.36 & 29.67& 16.24 & 31.81 & 10.08 & 5.23 & 20.32 & 19.09 & 25.31 & 25.88  & \textbf{30.15}     & 33.72\\ \midrule
    \multirow{3}{*}{$70\%$} & $100\%$ & $30\%$  & 17.45 & 27.14& 18.01 & 27.32 & 7.26 & 5.37 & 16.80 & 13.82 & 22.41 & 24.39 &   \textbf{27.21}   &  33.39\\ 
                             & $30\%$ & $100\%$ & 16.09 & 34.59 & 14.30 & 35.72 & 7.49 & 5.20 & 18.42 & 15.32 & 26.27 & 27.56  &  \textbf{31.03}    & 32.76\\ 
                             & $65\%$ & $65\%$  & 17.20 & 29.39 & 18.27 & 29.34 & 6.93 & 6.23 & 16.19 & 15.76 & 23.77 & 25.26 &   \textbf{28.97}   & 31.84 \\ \midrule
    \multirow{3}{*}{$90\%$} & $100\%$ & $10\%$  & 18.85 & 25.03 & 18.01 & 29.14 & 5.98 & 5.82 & 15.50 & 11.27 & 21.37 & 23.66 &   \textbf{26.11}   & 35.11\\ 
                             & $10\%$ & $100\%$ & 17.79 & 31.65 & 17.88 & 32.19 & 6.29 & 5.09 & 21.22 & 18.38 & 26.22 & 27.50 &  \textbf{30.23}    & 33.85 \\ 
                             & $55\%$ & $55\%$  & 16.28 & 27.89 & 16.20 & 31.47 & 6.84 & 6.39 & 16.82 & 8.44 & 25.42 & 25.82 &   \textbf{28.59}   &  32.79\\
    \bottomrule
    \end{tabular}    }
    % \captionsetup{belowskip=1pt}
    \caption{\textbf{Performance comparison on the {\datasettwo} dataset}. We report AP and FG  based on F1-Macro. Missing Modality can occur in both training and testing. The best performance is in \textbf{bold}.}
    \label{tab:mmimdb}
\end{table*}

\section{Experiments} 
\label{sec:exp}
\subsection{Datasets and Evaluation Metrics} 
Following the experiment protocol adopted in~\cite{zhao2024reconstruct}, we adapt two curated datasets established for the CMML task: {\datasetone} and {\datasettwo}. {\datasetone} benchmark is derived from the UPMC-Food101~\cite{wang2015recipe} dataset, comprising $101$ food categories and $61,142$ training, $6,846$ validation, and $22,716$ test image-text pairs. For the CL setup, these categories are partitioned into 10 sequential sessions. {\datasettwo} benchmark is built upon the MM-IMDb~\cite{arevalo2017gated} dataset. This dataset consists of $15,552$ training, $2,608$ validation, and $7,799$ test image-text pairs across 27 movie genres. Adhering to the multi-label continual learning paradigm~\cite{dong2023knowledge}, the 7 most infrequent genres are removed, and the remaining categories are structured into 5 distinct sessions, each introducing 4 new genres. For both benchmarks, the missing severity is controlled by a ratio $\eta\%$, representing the modality-incomplete data within each session. In cases where both modalities can be absent, this portion is typically split evenly between image-only and text-only data. 

For evaluation, we employ two standard metrics from the continual learning literature~\cite{chaudhry2018riemannian, chaudhry2018efficient, lopez2017gradient, loshchilov2017decoupled}: Average Performance~(AP) and Average Forgetting~(FG). AP is the average performance across all $T$ tasks after the model has completed training on the final task, calculated as $\mathrm{AP}=\frac{1}{T}\sum_{t=1}^{T}a_{t, T}$, where $a_{i,j}$ denotes the performance on task $\mathcal{T}_i$ after training up to task $\mathcal{T}_j$. FG measures the performance degradation on past tasks and is calculated as:
\begin{equation}
    \mathrm{FG} = \frac{1}{T-1}\sum_{t=1}^{T-1}\max_{z\in\{t,\cdots,T-1\}}(a_{t,z}-a_{t,T}).
\end{equation}
% $\mathrm{FG} = \frac{1}{T-1}\sum_{t=1}^{T-1}\max_{z\in\{t,\cdots,T-1\}}(a_{t,z}-a_{t,T})$.

\textbf{Comparison Methods.} To comprehensively evaluate our DeLo framework,  we benchmark it against a strong set of baselines from two relevant research domains: missing modality and continual learning. For the missing modality problem,  we include MAP~\cite{lee2023multimodal} and MSP~\cite{jang2024towards}), which represent modern methods that adapt pre-trained LMMs for modality-incomplete data. From the continual learning perspective, we compare against leading prompt-based frameworks to contrast with our LoRA-based approach, including L2P~\cite{wang2022learning} and DualPrompt~\cite{wang2022dualprompt}. Importantly, we also conduct a direct comparison with RebQ~\cite{zhao2024reconstruct}, the state-of-the-art method that first introduced the CMML task and its corresponding benchmark.

\textbf{Implementation Details.} Following~\cite{lee2023multimodal}, we utilize ViLT-B/32\footnote{https://huggingface.co/dandelin/vilt-b32-mlm}~\cite{kim2021vilt} as the pre-trained LMM backbone. Following the parameter-efficient paradigm, the parameters of the ViLT backbone remain frozen throughout all continual learning stages. A new, learnable fully-connected layer is instantiated as a dedicated classification head for each task. An empty string denotes the dummy text input~(i.e., $\tilde{\mathbf{t}}$), and an image with all pixel values equal one as the dummy image input~(i.e., $\tilde{\mathbf{v}}$). We set the expert pool size to $E=16$ and the selected rank to $r=4$. The loss hyperparameters are $\lambda_1=0.1$ and $\lambda_2=0.1$, and the EMA momentum is $\beta=0.99$. We train all model using the  AdamW~\cite{loshchilov2017decoupled} optimizer with a batch size of 4. The learning rate is initialized to $1$e-$4$ and managed by a cosine annealing scheduler with a linear warmup phase over the first $10\%$ of training steps. All experiments are conducted on one NVIDIA A100 GPU.

\subsection{Main Results}
Table~\ref{tab:food101} and Table~\ref{tab:mmimdb} summarizes the experimental results on both benchmarks, showing that our proposed DeLo framework consistently achieves the highest AP and maintains comparable or lower FG across all configurations. Unlike methods such as MAP and MSP which are not designed for continual learning, DeLo avoids the catastrophic forgetting that arises from overfitting adaptable parameters to each new task. Furthermore, DeLo improves upon advanced prompt-based continual learning methods. While approaches like L2P and DualPrompt utilize prompt pools, their performance can be hindered when missing modalities prevent effective pool access, our dynamic low-rank experts remain robust. Even RebQ, which extends this paradigm to CMML, still exhibits measurable forgetting in certain settings. The slightly higher FG of DeLo on the MM-IMDb-CMML can be attributed to its multi-label nature and richer semantic diversity, which make continual adaptation more challenging. We also observe that for all methods, performance correlates with the availability of text, likely due to the richer semantic information contained in textual descriptions.

% \begin{figure*}[t] 
%     \centering 
%     % \captionsetup{aboveskip=-34pt} 
%     \includegraphics[width=\textwidth]{AnonymousSubmission/Figure/modality-factor tSNE.pdf} 
%     \caption{The t-SNE visualization of learned modality-specific factor pools for each sequential task in \datasetone.}
%     \label{fig: modality facotr tSNE} 
% \end{figure*}

% The results on the more challenging multi-label MM-IMDb-CMML dataset (Table~\ref{tab:food101}) further validate DeLo's architectural advantages. Our method's impressive AP is directly anchored by its stability, as demonstrated by its low FG score. This performance confirms that our dual decomposition design effectively preserves knowledge regardless of task complexity, overcoming the interference limitations inherent in prior prompt-based approaches. The consistent performance gap underscores the benefits of DeLo’s design: a superior anti-forgetting architecture and the use of LoRA, a more powerful adaptation technique than prompt tuning. Finally, we note that across all methods, performance correlates with the availability of text, likely due to the rich semantic content of the movie plots in this benchmark.

\subsection{Ablation Study}
In this section, we explore various DeLo variants to assess the importance of each component. Experiments are all based on the {\datasetone} with a missing ratio $\eta=70\%$, where both image and text modalities are missing.

\subsubsection{Effectiveness of DeLo Components}
\begin{figure}[t] 
    \centering 
    \captionsetup{aboveskip=-2pt} 
    \includegraphics[width=\linewidth]{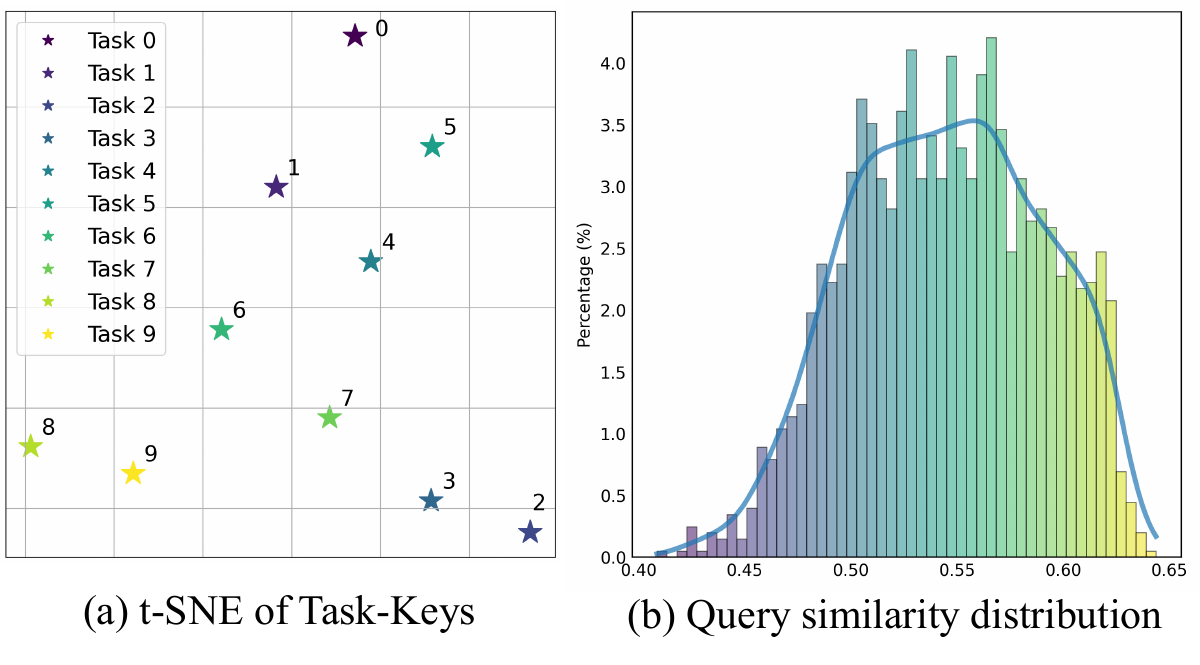} 
    \caption{(a) t-SNE visualization of the final representations learned by the Task-Key Memory for each of the 10 continual tasks in \datasetone. (b) The distribution of cosine similarity between visual and textual query signals, evaluated on modality-complete data.}
    \vspace{-1em}
    \label{fig:task-key and query similarity} 
\end{figure}

As presented in Table~\ref{tab:ablation one}, we compare our full DeLo model against several variants, each with a crucial component removed or simplified. The importance of our modality-decomposed architecture is evident when replacing the separate vision and text pools with a single unified pool (\texttt{w/o Modality-Specific Factor Pool}), which results in a significant performance drop due to modality interference. Performance collapses even more severely when removing the Cross-Modal Guided Routing mechanism (\texttt{w/o Cross-Modal Guide}), demonstrating that our strategy to preserve multimodal collaboration with incomplete data is critical. As illustrated in Figure~\ref{fig:task-key and query similarity}, the Task-Key Memory forms well-separated clusters for different continual tasks, while the visual and textual query signals exhibit strong semantic alignment. Additionally, replacing our dynamic decomposed LoRA experts with conventional static LoRA (\texttt{DeLo-LoRA}) also degrades performance. These results collectively confirm that each component of DeLo provides an essential contribution to its overall effectiveness.
\begin{table}[h] 
\captionsetup{belowskip=1pt}
    \centering
    \begin{tabular}{@{}lrr@{}}
    \toprule
    Method & AP~($\uparrow$) & FG~($\downarrow$) \\ \midrule
    DeLo ($E=16, r=4$) & \textbf{67.42} & \textbf{9.41} \\ \midrule
    DeLo-LoRA ($r=4$) & 60.39 & 12.08 \\
    w/o Cross-Modal Guide & 48.84 & 20.13 \\ 
    w/o Modality-Specific Factor Pool & 56.52 & 13.76 \\
    \bottomrule
    \end{tabular}
    \caption{Component-wise analysis of the DeLo framework.}
    \label{tab:ablation one}
\end{table}
\vspace{-2em}
\begin{table}[h] 
\captionsetup{belowskip=1pt}

    \centering
    \resizebox{\linewidth}{!}{
    \begin{tabular}{@{}cc|cccc@{}}
    \toprule
     \multirow{1}{*}{Factors} &  \multirow{1}{*}{Rank} &  \multirow{2}{*}{AP~($\uparrow$)} & \multirow{2}{*}{FG~($\downarrow$)} &  Trainable  & Total \\
     ($E$) &   ($r$) &  & &  Parameters & Parameters\\\midrule
    16 & 2 &  63.78 & 11.20 & 7.1M  & 182.8M \\
    16 & 4 &  67.42 & 9.41 & 12.4M  & 235.9M \\
    16 & 8 &  67.11 & 10.89 & 23.0M  & 342.1M \\
    32 & 2 &  64.53 & 13.37 & 14.2M  & 253.6M \\
    32 & 4 &  68.03 & 8.92 &24.8M  & 359.8M \\
    32 & 8 &  67.92 & 9.24 & 46.0M  & 572.1M \\
    \bottomrule
    \end{tabular}
    }
    \caption{Impact of factor pool size and selected rank on DeLo's performance and parameter efficiency.}
    \vspace{-1em}
    \label{tab:rank and factor}
\end{table}

\subsubsection{Impact of the Number of Rank and Factors} 
\begin{figure}[t] 
    \centering 
    \captionsetup{aboveskip=-2pt} 
    \includegraphics[width=0.96\linewidth]{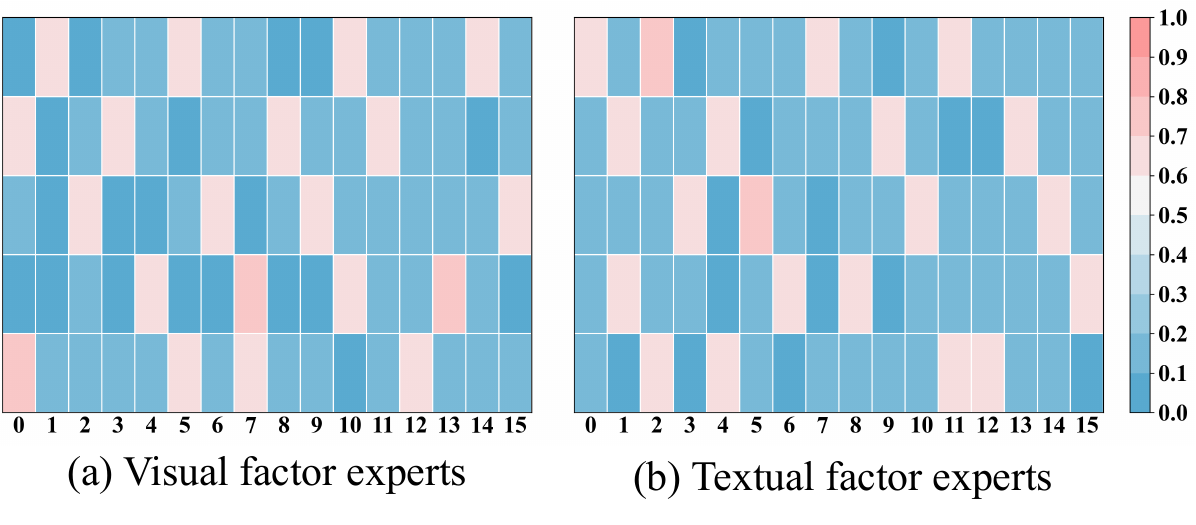} 
    \caption{Comparison of selection frequency of each factor expert for (a) the vision pool and (b) the text pool.}
    \vspace{-1em}
    \label{fig:Expert Utilization Heatmap} 
\end{figure}

We analyze the impact of the total number of available expert factors ($E$) and the selected rank ($r$) to determine an optimal balance between performance and parameter efficiency. As shown in Table~\ref{tab:rank and factor}, our results reveal a clear trend: increasing the rank from 2 to 4 yields a substantial improvement in AP, but this gain diminishes or slightly reverses when the rank is further increased to 8. This suggests that $r=4$ provides a sweet spot for expressive capacity without the potential optimization challenges of a larger rank. Meanwhile, a larger factor pool ($E=32$) consistently offers a modest performance advantage over a smaller one ($E=16$) by providing the router with a more diverse set of experts. Given that a core goal of our work is to maintain high parameter efficiency, we identify the setting of $E=16$ and $r=4$ as the optimal trade-off between predictive accuracy and computational cost. Figure~\ref{fig:Expert Utilization Heatmap} further shows the selection frequency of each factor expert from the visual and textual pools.

% We therefore adopt this setup as the default for our main experiments, with its trainable parameters constituting only about 5.3$\%$ of the total model size.

\subsubsection{Effectiveness of three loss components.}
Table~\ref{tab:ablation loss} demonstrates that our regularization terms are crucial for the effectiveness of the Cross-Modal Guided Routing mechanism. When training with only the standard classification loss ($\mathcal{L}_{\mathrm{c}}$), the model achieves a baseline performance. Without explicit constraints, the semantic alignment of the proxy queries used for missing modalities is not guaranteed, which limits the guidance mechanism's effectiveness. Introducing either the Alignment Loss ($\mathcal{L}_{\mathrm{align}}$) or the Consistency Loss ($\mathcal{L}_{\mathrm{con}}$) individually leads to a clear improvement in performance. The former enhances the proxy signal's quality by explicitly enforcing semantic similarity between queries, while the latter reinforces a consistent reasoning process by aligning the model's output distributions. Notably, the full model achieves the best results, suggesting that these two losses are complementary and work synergistically to enhance the reliability of our guided routing.
\begin{table}[h]
\captionsetup{belowskip=1pt}
\centering
\begin{tabular}{ccccc}
\toprule
$\mathcal{L}_{\mathrm{c}}$ & $\mathcal{L}_{\mathrm{align}}$ & $\mathcal{L}_{\mathrm{con}}$ & AP ($\uparrow$) & FG ($\downarrow$) \\
\midrule
\checkmark &  -  &   -       & 61.95 & 12.37 \\
\checkmark &   -       & \checkmark & 65.15 & 10.85 \\
\checkmark         & \checkmark & - & 65.88 & 10.74 \\
\midrule
\checkmark & \checkmark & \checkmark & \textbf{67.42} & \textbf{9.41} \\
\bottomrule
\end{tabular}
% \label{tab:ablation loss}
\caption{Impact of each loss component on performance.}
\vspace{-1em}
\label{tab:ablation loss}
\end{table}

\section{Conclusion}
In this paper, we propose DeLo, a novel framework that introduces a dual-decomposed low-rank expert architecture to the challenging problem of Continual Missing Modality Learning. DeLo dynamically constructs bespoke low-rank adaptations for each input by selecting factors from expert pools that are partitioned by both task and modality. This unique architecture, combined with a Cross-Modal Guided Routing strategy, enables knowledge preservation while robustly handling modality-incomplete data. Extensive experiments on two established CMML benchmarks demonstrate that DeLo significantly outperforms existing state-of-the-art methods, achieving superior average performance while showing robust knowledge retention across previous tasks.

\newpage

{
    % \bibstyle:{aaai2026}
    \bibliography{aaai2026}
}

\end{document}